\documentclass{article}

\PassOptionsToPackage{numbers, compress}{natbib}
\usepackage[final]{neurips_2022}

\usepackage[utf8]{inputenc} 
\usepackage[T1]{fontenc}    
\usepackage{hyperref}       
\usepackage{url}            
\usepackage{booktabs}       
\usepackage{amsfonts}       
\usepackage{nicefrac}       
\usepackage{microtype}      
\usepackage{xcolor}         
\usepackage{graphics}

\usepackage{multirow}
\usepackage{latexsym}
\usepackage{graphicx} 
\usepackage{algorithm}
\usepackage{algorithmic}
\usepackage{epsfig}
\usepackage{color}
\usepackage{amsmath}
\usepackage{graphicx,subfigure}
\usepackage{makecell}
\usepackage{textcomp}
\usepackage{url}
\usepackage{flushend}
\usepackage{ulem}
\usepackage{wrapfig}

\usepackage{listings}
\usepackage{color}

\definecolor{dkgreen}{rgb}{0,0.6,0}
\definecolor{gray}{rgb}{0.5,0.5,0.5}
\definecolor{mauve}{rgb}{0.58,0,0.82}

\vspace{-0.2in}

\title{Performance Law of Large Language Models}

\author{%
  Chuhan Wu  \\
  LLM Researcher \\
   \And Ruiming Tang \\
    LLM Researcher \\
}

\begin{document}

\maketitle

\vspace{-0.2in}

\begin{abstract}

\vspace{-0.1in}

Guided by the belief of the scaling law, large language models (LLMs) have achieved impressive performance in recent years.
However, scaling law only gives a qualitative estimation of loss, which is influenced by various factors such as model architectures, data distributions, tokenizers, and computation precision.
Thus, estimating the real performance of LLMs with different training settings rather than loss may be quite useful in practical development.
In this article, we present an empirical equation named ``Performance Law'' to directly predict the MMLU score of an LLM, which is a widely used metric to indicate the general capability of LLMs in real-world conversations and applications.
Based on only a few key hyperparameters of the LLM architecture and the size of training data, we obtain a quite accurate MMLU prediction of various LLMs with diverse sizes and architectures developed by different organizations in different years.
Performance law can be used to guide the choice of LLM architecture and the effective allocation of computational resources without extensive experiments.

\end{abstract}

\vspace{-0.15in}
\section{Introduction}
\vspace{-0.05in}

Owing to the foresight of the scaling law~\cite{kaplan2020scaling} and its derivations~\cite{hoffmann2022training}, the rapid development of large language models (LLMs) in recent years has generated a considerable impact on people's lifestyles and shaped many successful online applications~\cite{wu2023brief,owens2023nature}.
However, the huge computational overhead of LLMs presents significant trial-and-error expenses and uncertainty for their developers~\cite{touvron2023llama}, who may face increased pressure to optimize resource allocation and manage risks effectively.

To predict the performance of LLM before carrying out the whole experiment, researchers have proposed many variants of scaling laws~\cite{hoffmann2022training,krajewski2024scaling,sardana2023beyond,du2024understanding} to characterize the ability of LLM in different settings.
They are primarily focused on the training loss under a certain configuration.
However, the training loss scales of different models are diverse since they usually have distinct tokenizers and are optimized on different data recipes with various training strategies.
Several recent works on LLM performance prediction~\cite{ye2023predictable,owen2024predictable} are based on model sizes and training data.
However, they have limited accuracy and generality across different model structures (e.g., dense or sparse) and shapes (e.g., wide or deep), which may have substantial impacts on the final performance~\cite{kaplan2020scaling}.
Moreover, the precision and stability of the computing infrastructures are not taken into account, which usually damages the quality of the model~\cite{lee2024fp8}.
Thus, an accurate estimation of LLM performance is still missing for researchers to design and optimize their models to approach the ideal scaling law.

We discover an empirical equation to predict the MMLU~\cite{hendrycks2020measuring} metric of an LLM, which is a widely used measurement with good relevance to performance in downstream tasks~\cite{dubois2024length}.
We name this equation as ``Performance Law''.
All we need are the amount of training data and the key hyperparameters of a common Transformer-based LLM, including the number of layers, hidden size, and the intermediate size of feed-forward networks.
Even for mixture-of-expert (MoE) models, we only need to add the amount of activation parameters and follow a variant of dense model prediction.
By learning a few regression parameters of the equation on only \textbf{10} popular open source models released in \textbf{2024}, we obtain a surprisingly accurate performance prediction of LLMs of different sizes (from 0.5B to 1000+B) and in different years (from 2020 to 2024) released by different organizations around the world.
Performance law can help interpret various phenomena in real-world LLM development that are not fully covered by the scaling law.
It can guide the choice of LLM architecture under limited budgets to save computational resources and reduce the worrying carbon footprint of LLM training.

\section{Performance Law}

\subsection{Formulation of Performance Law}

For dense model, performance law involves the following key variables:
\begin{itemize}
\item The number of layers $N$
\item The hidden size $h$
\item The intermediate size $d$ of FFN 
\item The size of training data $T$ (trillion tokens)
\item The model size $S$ (billion parameters), which can be almostly derived from the above factors.
\end{itemize}
Motivated by the formulation of the scaling law, the capability of a model cannot increase unlimited on more and more data.
Thus, we introduce an empirical model saturation clip as follows:
$$T' = min(T, S)$$
where $T'$ represents the ``effective'' training tokens.
For example, give a model with 0.5B parameters, it may achieve quite satisfactory results by training on 500B tokens.

According to the scaling law, increasing the model size usually yields a lower final loss.
Unfortunately, this may not always be the case in practice.
For example, a large but slim model with 1,000 layers is likely to be highly unstable and it usually diverges in model training.
In addition, the performance of models with the same sizes varies under different model shapes according to our practice, which shows that model size does not largely decide model performance.
Inspired by~\cite{takase2023spike}, we devise a model unstable discount $u$ to characterize the instability of large model training:
$$ u = e^{-[(\frac{10}{d}+\frac{20}{h})(\gamma N)]^2}$$
where $\gamma$ reflects the precision loss of the training infrastructure (larger is worse) and we set $\gamma = 1$ by default.
The intuition of this equation is based on the ratio of high-activation parameters to the total number of parameters.
This formula implies that the performance of LLMs may be suboptimal if the model is too slim, which is consistent with existing experimental analysis.

Based on the preparation presented above, we propose the formulation of MMLU score prediction using a log-linear regression function as follows:
\begin{equation}
    \begin{aligned}
        \text{MMLU} &= w_1\log(uN) + w_2\log(uh) + w_3\log(ud) + w_4\log(uT') + b \\
                    &= w_1\log(N) + w_2\log(h) + w_3\log(d) + w_4\log(T') - \sum_{i=1}^4w_i(\frac{10}{d}+\frac{20}{h})(\gamma N)^2+ b
    \end{aligned}
\end{equation}
where $w_1, w_2, w_3, w_4$ and $b$ are regression parameters.
Since this simple function has only a few parameters, it is unlikely that it will overfit existing observations.
However, to maintain good generalization ability, we only use 10 open source models released in 2024 with sizes ranging from 7B to 405B, which are generally trained on high-quality data with carefully designed strategies.\footnote{It is easy to have higher regression accuracy by adding more models, but we prefer to show the surprising generalization ability of the simple prediction model in time and space with limited data samples in 2024.} Additionally, we up-sample the data points of LLMs with reported data contamination analysis to have a better prediction of the real generalization capabilities of LLMs.
After regression on these data points, we obtain the values of these learnable parameters: $w_1 = 13.95018, w_2 = 0.23072, w_3 = -0.48523, w_4 = 5.39802, b = 9.19541$.
Given a common dense model configuration, we can approximately predict its MMLU score across the entire training phase by varying the training tokens.

We then introduce the performance law of MoE models.
Existing performance prediction methods usually fail to estimate the results of MoE models due to their sparse activation mechanism.
Thus, we need to consider the number of activated parameters $A$ (billion parameters) in the performance prediction.
In our practice, we empirically find that the performance of MoEs with common shapes can be roughly regarded as the performance of a dense model with $\sqrt{AS}$ parameters.
For example, the performance of a 47B-MoE with 13B activated parameters is comparable with a 25B dense model.
However, it is not applicable in extreme cases where there are too few or too many activated parameters.
Therefore, we propose an expansion factor $g$ to comprehensively draw the capability gains of various MoE models as follows:
$$ g =(\frac{\sqrt{AS}}{A})^{\frac{1}{3}}\cdot \frac{0.5+\sqrt{\frac{A}{S}}}{1+e^{-\frac{A}{4}}}$$
where the first term describes the ``side length'' expansion of activated parameters and the second term adjusts the scale of this expansion according to the relative activation ratio and the absolute activation size.
We inherit the model unstable discount of the dense model but replace the dimension of dense FFN with the maximum FFN size $d'$ of activated experts, which is formulated as follows:
$$ u' = e^{-[(\frac{10}{d'}+\frac{20}{h})N]^2}$$
We have $d'=d$ in many situations, but $d'$ is much larger than $d$ in fine-grained MoE models such as DeepSeek-V2 and Qwen 57B-A14B.
For very small MoE models, we suggest adding this modified model saturation clip $T = min(T, \sqrt{AS})$.
We then predict the final MMLU performance by putting the expansion factor into the dense prediction function with the same parameters:
$$ \text{MMLU} = w_1\log(u'Ng) + w_2\log(u'hg) + w_3\log(u'd) + w_4\log(u'T) + b.$$

Note that since the upper bound of MMLU scores is 100, the prediction results cannot be unlimited.
Thus, to better estimate the performance of ultra-large models, we recommend using the following adjusting method for results greater than 90 to ensure that they are in reasonable ranges:
$$y = 90+10*\tanh(0.1x-9)$$
Note that the performance law can also be generalized to other benchmarks that comprehensively measure model capabilities.
For example, for strong models ($\text{MMLU} > 70$) we can obtain an approximate estimation of the MMLU-Pro~\cite{wang2024mmlu} performance by using a simple linear mapping $y = 2.33x-133$ obtained by linear regression, which means that an MMLU score of 82 corresponds to 72.04 on MMLU-Pro.
We also see a very high correlation between BIG-Bench~\cite{srivastava2022beyond} and MMLU scores, thereby developers can map their results easily.

\subsection{Examples of Performance Prediction}
Here we directly present the Python code for performance law prediction on a dense model and an MoE with common settings.

\begin{lstlisting}
# 7B Dense with 3T Training Tokens
n_layer = 32
hidden = 4096
ffn = 14336
token_T = 3
model_size = 7
token_T = min(token_T, model_size)
unstable_discount = np.exp(-((10/ffn+20/hidden)*n_layer)**2)
print(np.sum(np.array([13.95018, 0.23072, -0.48523,  5.39802]) * np.log(unstable_discount * np.array([n_layer,hidden,ffn,token_T]))) + 9.19541)
# Output: 60.13969302998589

\end{lstlisting}

\begin{lstlisting}
# 141B A39 MoE with 10T Training Tokens
total = 141
act = 39
ratio = (np.sqrt(act*total)/act)**(1/3)*((1+np.sqrt(4*act/total))/2)*(1/(1+np.exp(-act/4)))
n_layer = 56 * ratio
hidden = 6144 * ratio
ffn = 16384 
token_T = 10
model_size = np.sqrt(act*total)
token_T = min(token_T, model_size)
unstable_discount = np.exp(-((10/ffn+20/hidden)*(n_layer))**2)
print(np.sum(np.array([13.95018, 0.23072, -0.48523,  5.39802]) * np.log(unstable_discount * np.array([n_layer,hidden,ffn,token_T]))) + 9.19541)
# Output: 77.50985935370231

\end{lstlisting}

\begin{figure}[h]
    \centering
    \subfigure[All models.]{
    \includegraphics[width=0.48\linewidth]{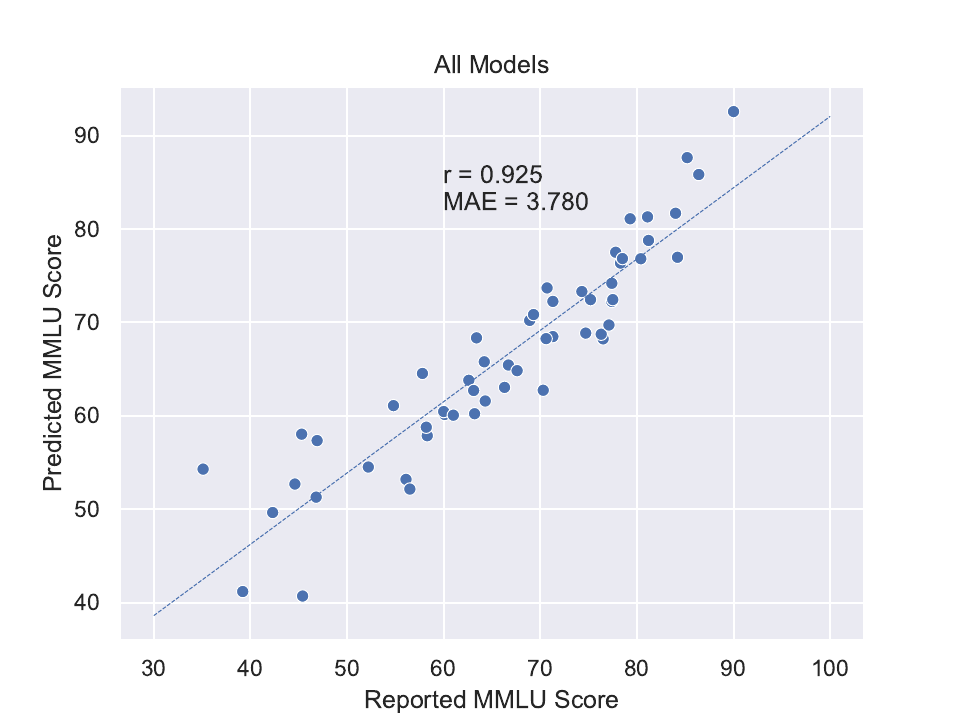}
    }
        \subfigure[English models (excluded Llama1).]{
    \includegraphics[width=0.48\linewidth]{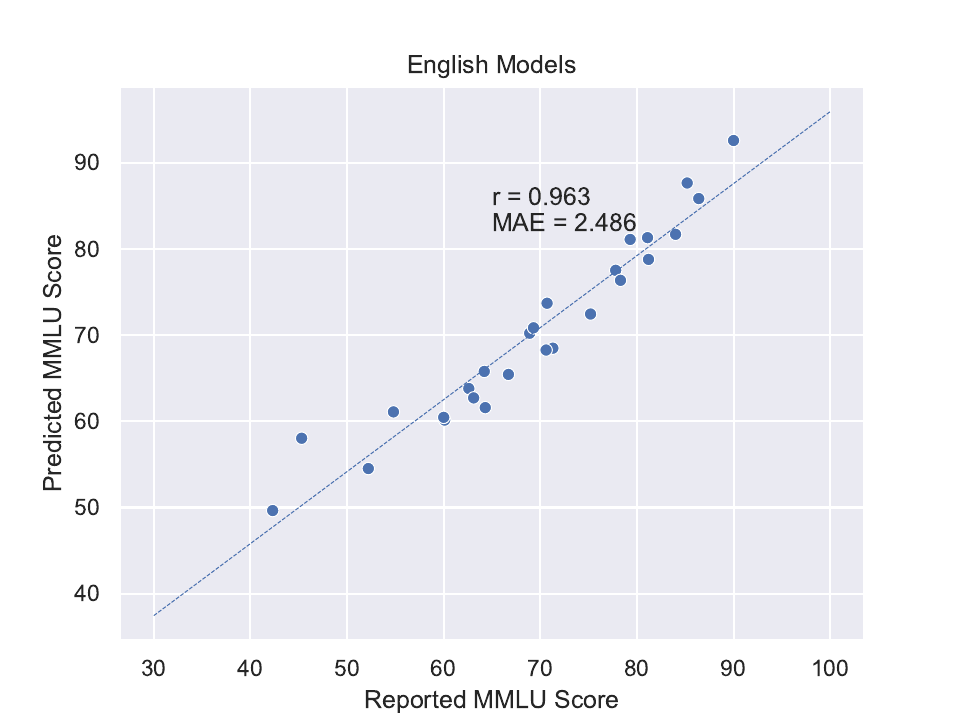}
    }
    \caption{The prediction and real MMLU scores of different models.}
    \label{fig1}
\end{figure}

\subsection{Results and Insights}

The results of predicted and reported results of 55 models are summarized in Fig.~\ref{fig1} and Table~\ref{table1}.
We can see that performance law has quite accurate predictions over all types of models ranging from 0.5B to 1000+B, especially on English-oriented models trained with high-quality data.
The results demonstrate good precision and generality of the performance law of large language models.

\begin{table}[h]
\caption{An overview of performance law predictions for various LLMs. *Some unknown values are random guesses of the authors. They are probably inaccurate and are just shown for prediction.}\label{table1}
\centering
\resizebox{\linewidth}{!}{
\begin{tabular}{lccccccccccc}
\hline
Model                    & Layer & Hidden & FFN   & \begin{tabular}[c]{c@{}}Expert\\ FFN $d'$\end{tabular} & Token/T & Size/B & Act./B & MoE & MMLU & \multicolumn{1}{c}{Prediction} & \multicolumn{1}{c}{Diff} \\ \hline
Llama 7B                 & 32    & 4096   & 11008 & -                                                           & 1       & 7      & -            & -   & 35.1 & 54.29                          & -19.19                   \\
Llama 13B                & 40    & 5120   & 13824 & -                                                           & 1       & 13     & -            & -   & 46.9 & 57.35                          & -10.45                   \\
Llama 33B                & 60    & 6656   & 17920 & -                                                           & 1.4     & 33     & -            & -   & 57.8 & 64.53                          & -6.73                    \\
Llama 65B                & 80    & 8192   & 22016 & -                                                           & 1.4     & 70     & -            & -   & 63.4 & 68.34                          & -4.94                    \\
Llama2 7B                & 32    & 4096   & 11008 & -                                                           & 2       & 7      & -            & -   & 45.3 & 58.03                          & -12.73                   \\
Llama2 13B               & 40    & 5120   & 13824 & -                                                           & 2       & 13     & -            & -   & 54.8 & 61.09                          & -6.29                    \\
Llama2 34B               & 48    & 8192   & 22016 & -                                                           & 2       & 34     & -            & -   & 62.6 & 63.80                          & -1.20                    \\
Llama2 70B               & 80    & 8192   & 28672 & -                                                           & 2       & 70     & -            & -   & 68.9 & 70.21                          & -1.31                    \\
Llama3.1 8B              & 32    & 4096   & 14336 & -                                                           & 15      & 8      & -            & -   & 66.7 & 65.43                          & 1.27                     \\
Llama3.1 70B             & 80    & 8192   & 28672 & -                                                           & 15      & 70     & -            & -   & 79.3 & 81.09                          & -1.79                    \\
Llama3.1 405B            & 126   & 16384  & 53248 & -                                                           & 15      & 405    & -            & -   & 85.2 & 87.64                          & -2.44                    \\
Gemma 2B                 & 18    & 2048   & 16384 & -                                                           & 3       & 2      & -            & -   & 42.3 & 49.64                          & -7.34                    \\
Gemma 7B                 & 28    & 3072   & 24576 & -                                                           & 6       & 7      & -            & -   & 64.3 & 61.58                          & 2.72                     \\
Gemma2 2B                & 26    & 2304   & 9216  & -                                                           & 2       & 2      & -            & -   & 52.2 & 54.51                          & -2.31                    \\
Gemma2 9B                & 42    & 3584   & 14336 & -                                                           & 8       & 9      & -            & -   & 71.3 & 68.48                          & 2.82                     \\
Gemma2 27B               & 46    & 4608   & 36864 & -                                                           & 13      & 27     & -            & -   & 75.2 & 72.44                          & 2.76                     \\
Mistral 7B               & 32    & 4096   & 14336 & -                                                           & 3       & 7      & -            & -   & 60.1 & 60.14                          & -0.04                    \\
Mixtral 8*7B             & 32    & 4096   & 14336 & 14336                                                       & 8       & 47     & 13           & $\checkmark$   & 70.6 & 68.26                          & 2.34                     \\
Mixtral 8*22B            & 56    & 6144   & 16384 & 16384                                                       & \sout{10}*      & 141    & 39           & $\checkmark$   & 77.8 & 77.51                          & 0.29                     \\
Mistral Large            & 88    & 12288  & 28672 & -                                                           & \sout{7}*       & 123    & -            & -   & 81.2 & 78.77                          & 2.43                     \\
Mistral Large 2          & 88    & 12288  & 28672 & -                                                           & \sout{12}*      & 123    & -            & -   & 84.0   & 81.68                          & 2.32                     \\
PaLM 540B                & 118   & 18432  & 73728 & -                                                           & 0.78    & 540    & -            & -   & 69.3 & 70.84                          & -1.54                    \\
Falcon 180B              & 80    & 14848  & 59392 & -                                                           & 3.5     & 180    & -            & -   & 70.7 & 73.69                          & -2.99                    \\
Text-davinci-002         & 96    & 12288  & 49152 & -                                                           & 0.3     & 175    & -            & -   & 63.1 & 62.71                          & 0.39                     \\
GPT-4                    & \sout{96}*    & \sout{12288}*  & \sout{49152}* & \sout{49152}*                                                       & 13      & 1831   & 280          & $\checkmark$   & 86.4 & 85.83                          & 0.57                     \\
PaLM2 340B               & \sout{96}*    & \sout{18432}*  & \sout{73728}* & -                                                           & 3.6     & 340    & -            & -   & 78.3 & 76.35                          & 1.95                     \\
Gemini Ultra $\sim$1760B & \sout{200}*   & \sout{32768}*  & \sout{65536}* & -                                                           & 11      & 1760   & -            & -   & 90.0   & 92.57                          & -2.57                    \\
Nemotron 15B             & 32    & 6144   & \sout{16384}* & -                                                           & 8       & 15     & -            & -   & 64.2 & 65.78                          & -1.58                    \\
Nemotron 340B            & 96    & 18432  & 73728 & -                                                           & 9       & 340    & -            & -   & 81.1 & 81.29                          & -0.19                    \\
Gopher 280B              & 80    & 16384  & 65536 & -                                                           & 0.3     & 280    & -            & -   & 60   & 60.45                          & -0.45                    \\
Deepseek-67B             & 95    & 8192   & 22016 & -                                                           & 2       & 67     & -            & -   & 71.3 & 72.25                          & -0.95                    \\
Deepseek-V2              & 60    & 5120   & 1536  & 12288                                                       & 8.1     & 236    & 21           & $\checkmark$   & 78.5 & 76.83                          & 1.67                     \\
DeekSeek-V2-Lite         & 27    & 2048   & 1408  & 10944                                                       & 5.7     & 16     & 2.4          & $\checkmark$   & 58.3 & 57.86                          & 0.44                     \\
Skywork-MoE              & 52    & 4608   & 12288 & 12288                                                       & \sout{8}*       & 146    & 22           & $\checkmark$   & 77.4 & 74.19                          & 3.21                     \\
Qwen 1.8B                & 24    & 2048   & 11008 & -                                                           & 2       & 1.8    & -            & -   & 44.6 & 52.69                          & -8.09                    \\
Qwen 7B                  & 32    & 4096   & 22016 & -                                                           & 2.4     & 7      & -            & -   & 58.2 & 58.78                          & -0.58                    \\
Qwen 14B                 & 40    & 5120   & 27392 & -                                                           & 3       & 14     & -            & -   & 66.3 & 63.04                          & 3.26                     \\
Qwen 72B                 & 80    & 8192   & 49152 & -                                                           & 3       & 72     & -            & -   & 77.4 & 72.24                          & 5.16                     \\
Qwen 1.5 0.5B            & 21    & 1024   & 2816  & -                                                           & 2.4     & 0.5    & -            & -   & 39.2 & 41.18                          & -1.98                    \\
Qwen 1.5 1.8B            & 21    & 2048   & 5504  & -                                                           & 2.4     & 1.8    & -            & -   & 46.8 & 51.29                          & -4.49                    \\
Qwen 1.5 4B              & 21    & 2560   & 6912  & -                                                           & 2.4     & 4      & -            & -   & 56.1 & 53.19                          & 2.91                     \\
Qwen 1.5 7B              & 28    & 4096   & 11008 & -                                                           & 4       & 7      & -            & -   & 61.0   & 60.06                          & 0.94                     \\
Qwen 1.5 14B             & 40    & 5120   & 13696 & -                                                           & 4       & 14     & -            & -   & 67.6 & 64.83                          & 2.77                     \\
Qwen 1.5 32B             & 64    & 5120   & 27392 & -                                                           & \sout{7}*       & 32     & -            & -   & 74.3 & 73.30                          & 1.00                     \\
Qwen 1.5 72B             & 80    & 8192   & 24576 & -                                                           & 3       & 72     & -            & -   & 77.5 & 72.44                          & 5.06                     \\
Qwen 1.5 110B            & 80    & 8192   & 49152 & -                                                           & \sout{7}*       & 110    & -            & -   & 80.4 & 76.81                          & 3.59                     \\
Qwen 2 0.5B              & 24    & 896    & 4864  & -                                                           & 12      & 0.5    & -            & -   & 45.4 & 40.70                          & 4.70                     \\
Qwen 2 1.5B              & 28    & 1536   & 8960  & -                                                           & 7       & 1.5    & -            & -   & 56.5 & 52.15                          & 4.35                     \\
Qwen 2 7B                & 28    & 3584   & 18944 & -                                                           & 7       & 7      & -            & -   & 70.3 & 62.73                          & 7.57                     \\
Qwen 2 72B               & 80    & 8192   & 29568 & -                                                           & 7       & 72     & -            & -   & 84.2 & 76.97                          & 7.23                     \\
Qwen 2 57B-A14B          & 28    & 3584   & 18944 & 20480                                                       & 11.5    & 57     & 14           & $\checkmark$   & 76.5 & 68.24                          & 8.26                     \\
Yi 6B                    & 32    & 4096   & 11008 & -                                                           & 3       & 6      & -            & -   & 63.2 & 60.22                          & 2.98                     \\
Yi 34B                   & 60    & 7168   & 20480 & -                                                           & 3       & 34     & -            & -   & 76.3 & 68.73                          & 7.57                     \\
Yi-1.5 34B               & 60    & 7168   & 20480 & -                                                           & 3.6     & 34     & -            & -   & 77.1 & 69.72                          & 7.38                     \\
GLM-4 9B                 & 40    & 4096   & 13696 & -                                                           & 10      & 9      & -            & -   & 74.7 & 68.85                          & 5.85 \\  
\hline                 
\end{tabular}
}
\vspace{0.1in}
\end{table}

We can derive many insights from the implication and prediction of the performance law.
Here we summarize a few critical ones according to our experience.

\textbf{Model depth is a double-edged sword.}

According to the regression coefficient, the number of layers has the most positive impact on the model performance.
It is intuitive because increasing model depth can rapidly introduce more non-linearity.
This also explains the phenomenon that large dense models without sufficient depth (e.g., a 96-layer huge model) may not have a dramatic advantage over an 80-layer 70B model.
However, different from the ideal prediction of the scaling law, in practice layer depth is not everything.
The instability of deep model training hinders developers from increasing the model depth.
If the training infrastructure has computation precision issues (this is unfortunately common), the quality of deep models would be heavily damaged.

\begin{figure}[t]
    \centering 
    \includegraphics[width=0.7\linewidth]{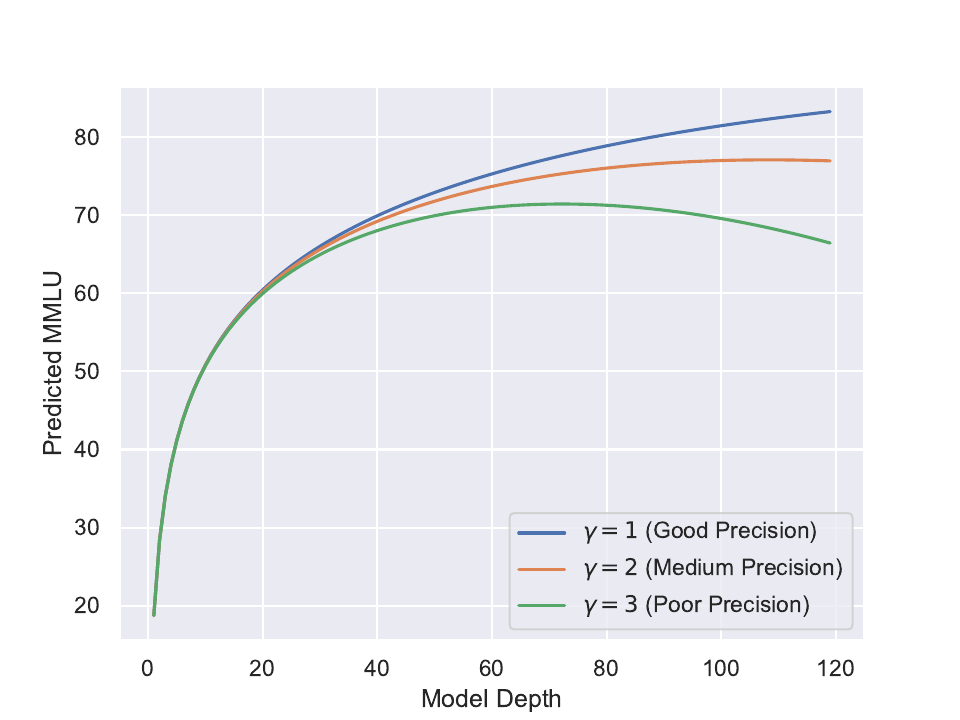}
  
    \caption{The predicted MMLU scores under different model depths and precision degrees.}
    \label{fig2}
\end{figure}

Fig.~\ref{fig2} shows the predicted performance under different precision coefficients $\gamma$.
Here we set the hidden dimension to 8192 and the FFN size to 28672.
When the training infrastructure has good precision ($\gamma=1$, can be achieved by leading companies), it is possible to train models with more than 100 layers without affecting their performance.
Unfortunately, the predicted performance decays rapidly if the precision is not satisfactory, which is consistent with common practical experience in the industry.
Moreover, given the same size, deep models usually have lower inference speeds and larger consumption of KV cache memory.
Thus, if developers are concerned more about inference efficiency or are not confident about the training stability and precision, it is recommended to use a relatively shallow model structure to reduce the negative effect of bad precision.

\textbf{Hidden size is (sometimes) more important than the FFN size.}

If the model is trained from scratch rather than expanded from another one, it may be better to use a relatively larger hidden size rather than the FFN size, as shown by the regression coefficients in the performance law.
This may be due to the high redundancy of parameters in the large FFN matrices.
However, the performance does \textbf{NOT} have a simple negative correlation with the FFN size, since it contributes to controlling the unstable discount $u$.
For example, the ratios of FFN size to hidden size in Llama 405B (3.25) and Mistral Large (2.33) are not too large.
This is quite consistent with the experimental results of the scaling law paper~\cite{kaplan2020scaling} and our own practice.
However, increasing the size of FFN may be a compromise due to the limitation of the inference cost and the precision of model training.
Thus, choosing a moderate FFN size may be feasible for LLM designers.

\textbf{Mainstream attention architectures seem to have similar performance.}

There are many different variants of standard attention architectures, such as GQA~\cite{ainslie2023gqa} and MLA~\cite{deepseekv2}.
The design of FFN also changes from a bottleneck network to GLU~\cite{shazeer2020glu}.
However, we hypothesize that they may have similar performance given the accurate prediction of performance law, though they may have small performance gaps in practice.
For example, according to the Gemma 2 report~\cite{team2024gemma2}, GQA may have a performance similar to that of MHA.
This shows that we still do not find a ``shortcut'' to greatly improve the performance of Transformer-based models under the same but sufficient computing power.

\textbf{Top players have similar data quality.}

Data quality and distributions usually largely affect model quality.
However, the prediction accuracy is still satisfactory even if data quality and distribution are not considered by the performance law.
This implies a reason that top LLM players may have similar data qualities and distributions (although the proportion of certain languages may be increased).
This could be reasonable because high-quality data sources, such as books, papers, news articles, exam exercises, and code repositories, are commonly used by all mainstream LLMs.
Some common data filtering and processing techniques are also shared.
These reasons cause the inherent homogeneity of the data used by different organizations.
Some leading organizations may have significantly more and higher quality data than their competitors.
This would make it possible to exceed the prediction of the performance law.

In another view, we need to acknowledge the influence of knowledge-intensive data such as exam problems.
These forms of data are more relevant to MMLU evaluation than common webpages and books, and training more on these knowledge-intensive data may lead to an increase in MMLU scores.
This explains why the performance of the first-generation Llama is far below prediction.
However, we can see that larger models may narrow this gap, as demonstrated by Llama 65B, Text-Davinci, Gopher, and PaLM.
In addition, we can see that most Chinese-enhanced models perform better than predictions.
This is probably because the proportion of exam and exercise data is significantly higher in Chinese model data mixtures.

\textbf{MoE is promising but hard to train.}

According to our analysis, MoE models can achieve comparable performance with a larger dense model with less training and inference cost.
Furthermore, fine-grained MoE such as Deepseek V2 and Qwen 57B-A14B may have a slight advantage over the coarse-grained MoE due to the shrinking of FFN size.
Based on our performance law, designers may be able to flexibly adjust the configurations of MoE models based on their requirements.
However, the number of successful cases of MoE-based LLM is somewhat limited, and the training MFU of MoE is usually much worse than the dense model.
In another view, MoE can be upcycled or transformed from an off-the-shelf dense model without training from scratch~\cite{llama-moe}.
Thus, developers may need to be prudent when making the decision between MoE and dense models with limited budgets.

\section{Applications and Implications}

Performance law has many applications in the development of LLM by changing different variables in its formulation.
Here we list a few possible ones based on our experience.

\subsection{Predict the Upscaling Potential of LLMs}

\begin{figure}[t]
    \centering 
    \includegraphics[width=0.7\linewidth]{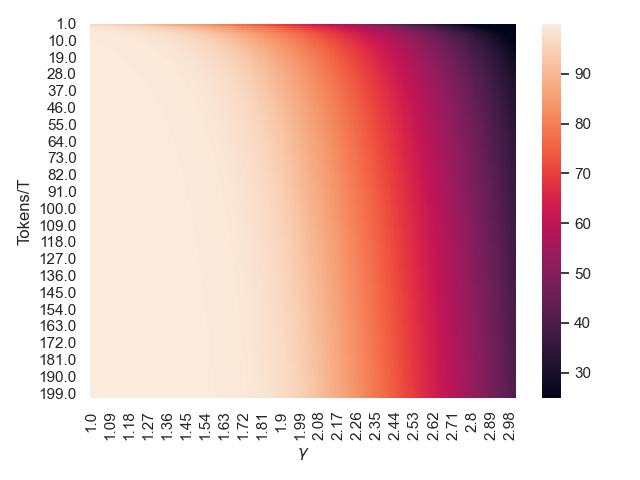}
  
    \caption{Predictions under different data sizes and $\gamma$ values.}
    \label{fig3}
\end{figure}

Performance law can be used to predict the potential of the next generation LLMs.
Just imagine a giant MoE model with 125T parameters in total and 22T activated parameters, which are much more than any existing models.
We set an extreme case to demonstrate our prediction, where the model has 1300 layers, a hidden size of 51200, and an FFN size of 65536.
It is probably too difficult to achieve $\gamma=1$ since the model is too deep, thus we vary the value of $\gamma$ from 1 to 3 and show the model performance under different numbers of training tokens in Fig.~\ref{fig3}.
We find that the value of $\gamma$ must be smaller than 2 if the model performance is expected to be stronger than the current SOTA.
This would be a tough challenge given such a deep model.
If we set $\gamma=1.9$ and the data size to 100T tokens, the final prediction result is 94.77, which may show a substantial intelligence gap over existing models.
Thus, more advanced model optimization strategies and training infrastructures are extremely required to ensure impressive performance of future giant models.

\subsection{Design Proper Model Architectures}

Another direct application of the performance law is helping design model architectures.
Given a specific training cost budget and inference efficiency requirements, developers can use our formula to search for suitable model architectures.
They can also customize the prediction coefficients based on their existing experiments to achieve higher prediction accuracy.
In addition, developers can adjust the value of $\gamma$ if they have a greater preference for inference speed or are less confident about their training strategies and infrastructures (they can estimate the value of $\gamma$ based on historical observations).
By planning the experiments using performance law as a reference, we hope that LLM researchers can greatly save the cost of heavy experiments.
We provide an easy-to-use tool to calculate possible architectures using a Python GUI program at \url{https://github.com/wuch15/performance-law-planner}.

\subsection{Tracking the Health Status of Models}

\begin{figure}[t]
    \centering
    \subfigure[Healthy model.]{
    \includegraphics[width=0.48\linewidth]{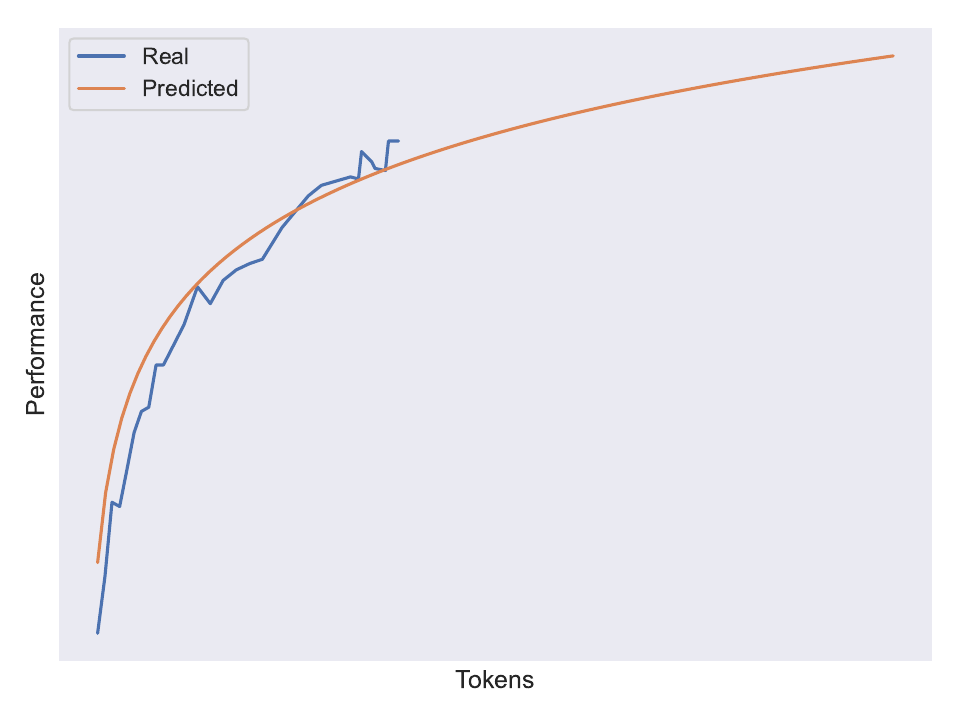}
    }
        \subfigure[Unhealthy model.]{
    \includegraphics[width=0.48\linewidth]{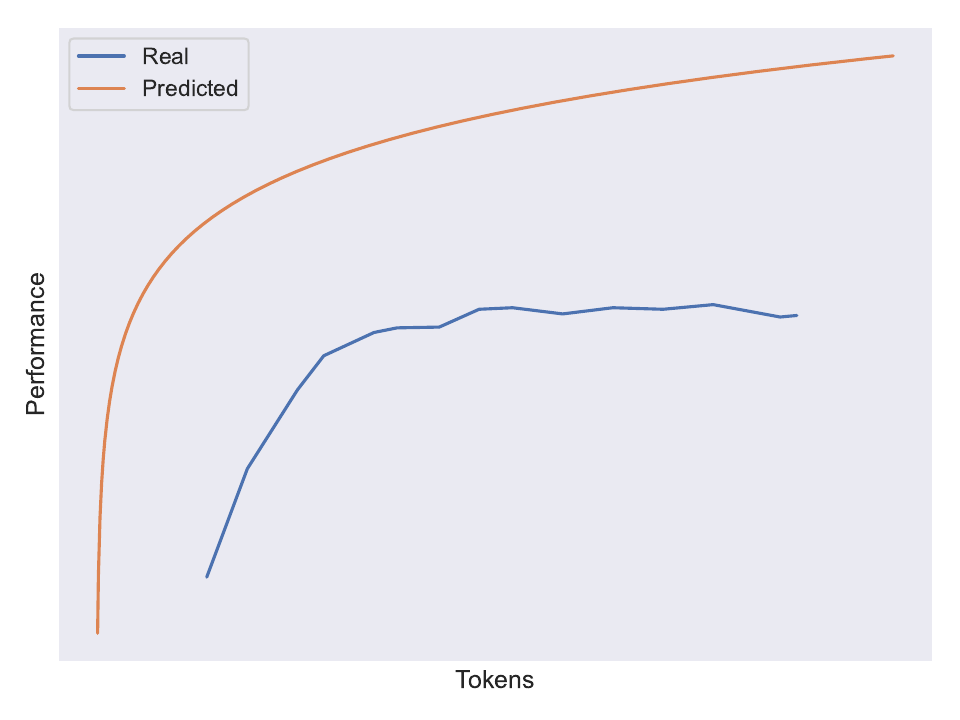}
    }
    \caption{The prediction and real metrics of two models.}
    \label{fig4}
\end{figure}

Performance law can also help track whether a model is in good status. 
Fig.~\ref{fig4} shows the predicted and real values of two models in our experiments.
We can assert that the first model is probably healthy since it well matches the prediction of performance law.
However, the result of the second model is abnormal and we can find this exception early without incorporating more sunk cost (e.g., the inferred $\gamma$ of the right figure is about 3.7, which is far above expectation; we find this is due to several implementation bugs after comprehensive examinations).
This may help LLM developers be aware of potential problems in their data, training strategies, and computation precision and fix possible bugs earlier before causing heavier losses.

\subsection{Planning Dense Model Expansion}

An interesting and practical application of the performance law is to predict the performance gain of model expansion techniques.
This is a common setting when LLM developers want to reuse a well-optimized small model and expand it into a larger one so that the cost of developing the small one can be further exploited.
We assume that the small dense model has $N_1$ layers and its hidden and FFN sizes are $h_1$ and $d_1$, which is optimized on $T_1$ trillions of tokens.
The target large model has $N_2$ layers with $h_1$ hidden dimensions and  $d_2$ FFN dimensions.
When using proper model expansion methods and train the obtained large model on $T_2$ trillions of tokens, its performance can be estimated using the example code as follows:

\begin{lstlisting}
# Train a 7B model on 3T tokens, then expand it to 70B and train 1T tokens
S_1, T_1, N_1, h_1, d_1= 7, 3, 32, 4096, 14336
S_2, T_2, N_2, h_2, d_2= 70, 1, 80, 8192, 28672
ratio = ((S_1*T_1)+(S_2*T_2))/(T_1+T_2)/S_2 - T_1*S_1/S_2/(1+np.exp((T_2)/0.1)) 
n_layer = N_1+(N_2-N_1)*ratio
hidden = h_1+(h_2-h_1)*ratio
ffn = d_1+(d_2-d_1)*ratio
    
unstable_discount = np.exp(-((10/ffn+20/hidden)*n_layer)**2)
print(np.sum(np.array([13.95018, 0.23072, -0.48523,  5.39802]) * np.log(unstable_discount * np.array([n_layer,hidden,ffn,T_1+T_2]))) + 9.19541)
# Output: 67.00187378584985

\end{lstlisting}

We find the predicted performance is quite consistent with our practical experience at different scales.
For well-optimized small models, expanding it to a larger one usually needs some training data to recover from the damage of the expansion operation, which is reflected by the second term of the "ratio" variable.
Based on this prediction, there usually exist some critical points (not too early or too late) to start the model expansion when the total budget limitation is given.
Thus, this application of the performance law can serve as a tool for developers to better plan and allocate their computing resources when they have to upcycle a large dense model from a smaller one.

\subsection{Check Implicit Data Contamination}

Data contamination is a common issue in LLM development.
The training and even test sets of many evaluation benchmarks exist in various data sources and might be occasionally contained by the training data.
Thus, if the tested model performance is abnormally higher than the prediction of the performance law (e.g., the prediction is 60 but the real score is 80), there may be a potential risk of data contamination, and a careful examination of training data is recommended.

\subsection{Infer the Model Structure and Data Size of Closed-Source Models}

Many organizations do not disclose all the details about their closed-source models.
However, it is possible to roughly guess the model architecture and data size based on reported MMLU scores and partial configuration information.
Note that the prediction needs to be adjusted based on the assessment of their data quality.

\section{Discussions}

\subsection{Reasons Behind Prediction Errors}

Although the prediction is reasonable in many cases, it is far from characterizing the real practice of LLM training.
Here we present several reasons for the errors of predictions.
Possible reasons for overestimation are as follows:

\begin{center}
\fcolorbox{black}{gray!10}{\parbox{1.\linewidth}{
\textbf{- The quality and distribution of training data are suboptimal.} 

Developers may consider improving the quality of their data or adjusting the recipe of the data mixture, e.g., by adding more knowledge-intensive data like exam questions. 

\vspace{0.1in}

\textbf{- The vocabulary does not well fit the data and model size.} 

The quality and size of vocabulary usually have some impact on the performance.
If the vocabulary does not match the data well, the model may tend to learn the shortcut of word composition rather than encoded knowledge.
In addition, if the vocabulary size is too large for a small model, it would be better to perform input-output parameter sharing to ensure that there are sufficient non-embedding parameters.
Larger models may also consider using larger vocabularies to enhance the data encoding efficiency in model learning~\cite{tao2024scaling}.

\vspace{0.1in}

\textbf{- The hardware and model training framework have bugs or precision issues.} 

This will significantly increase the instability discount in the performance law, especially for large and deep models. Finding and fixing these problems would be a quite painful but necessary experience for LLM developers.  
\vspace{0.1in}

\textbf{- The training hyperparameters are not well-tuned.} 

There are various factors that affect the final performance of LLMs, such as the initialization, learning rate scheduler, hyperparameters of the AdamW optimizer, and multi-stage training strategies. Finding appropriate hyperparameters for target training data and model architecture would be helpful.

}}
\end{center}

Several possible reasons for underestimation are as follows:

\begin{center}
\fcolorbox{black}{gray!10}{\parbox{1.\linewidth}{

\textbf{- The quality and recipe of training data are superior.} 

Leading companies usually have higher-quality data than other organizations, and the data mixture is better optimized based on solid experiments. 
\vspace{0.1in}

 \textbf{- The model architecture is more effective than Transformer.} 
 
 Performance may be better in the same configurations if the model has substantial differences from the standard Transformer.
\vspace{0.1in}
 
 \textbf{- The training strategy is quite effective.} 
 
 The final performance may be higher if the model is trained in ways with high learning efficiency and robustness, e.g., warm-start initialization or special learning-rate scheduler.
\vspace{0.1in}
 
 \textbf{- The model is trained on more MMLU-relevant data.} 
 
 Even after contamination checking, the model performance on MMLU may still be higher than expectations if trained on too much exam-like data. This effect is further amplified by annealing-like training strategies.

}}
\end{center}

\subsection{Are We Really Making Much Progress in These Years?}

In the era of GPT-3, the models with 100+ billion parameters are only trained on no more than 1 trillion tokens.
However, their performance can still be accurately predicted by the performance law that fits the data points only in 2024.
Based on the surprising generalization ability of performance law across years, we raise a tough question: Are we making much progress in these years?

It is hard to fully answer this challenge, but at least a major progress made by the community is the accumulation of a larger amount of high-quality data.
Picking 1T tokens of high-quality data is not difficult for common LLM players, even if they only process and filter publicly released datasets.
However, obtaining more than 10T tokens of data with the same quality standard becomes much more challenging because most data usually come from webpages and only a small fraction of them are valuable for model training.
Furthermore, human-generated high-quality data such as books and papers are vanishing~\cite{wild2024millions}, while some low-quality AI-generated content is polluting the Web and even academic publications.
This problem would be rather severe for building next-generation models with higher demands of data quality and amount.

Another notable progress is in building the ecosystem for the development and applications of LLM.
Without the collaborative effort of the community around the world, the hardware and software for LLM training and deployment cannot rapidly achieve maturity.
This may greatly reduce the risks of encountering problems that heavily affect the final model performance.
However, we are still facing great challenges in training very large language models.
This would explain why the prediction of performance law still fits large models in earlier years.

Doubts about the upscaling potential of LLMs continue, and the contest on relatively smaller models is heating up due to the pressure of business return and feasibility in downstream applications.
However, we still hope to see the future picture of LLMs at an unprecedented scale, to see whether the quantitative change of computational power makes a qualitative difference to machine intelligence.
\section{Conclusion}

In this paper, we present a performance law of LLMs that accurately predicts their performance on MMLU evaluation.
We find a surprising correlation between real results and our prediction over released years, model sizes, and developers.
We further derive several conclusions based on our prediction and provide rich analysis and interpretation from the perspective of industrial LLM development.
Using performance law as a simple reference, developers can better design and monitor their LLMs to save the computation overhead of experiments and reduce carbon dioxide emissions during model training.
We also hope the performance law can give a glimpse into the charming picture of the next generation of large language models.

\bibliographystyle{ACM-Reference-Format}
\bibliography{main}


 \appendix

 \section{Appendix}

Cases of models after the released date of this work:
\begin{itemize}
    \item OLMoE~\cite{muennighoff2024olmoe}: Real score 54.1, predicted score 56.21.
     \item MiniCPM3~\footnote{https://huggingface.co/openbmb/MiniCPM3-4B}: Real score 67.2, predicted score 65.36.   
\end{itemize}

We can see that performance law has good generalization ability on ``future'' models.

\end{document}